\pdfoutput=1
\PassOptionsToPackage{table}{xcolor}
\documentclass{article}
\usepackage{iclr2027_conference,times}
\usepackage[T1]{fontenc}
\usepackage[utf8]{inputenc}
\usepackage{amsmath,amsfonts,amssymb,mathtools,bm}
\usepackage{graphicx,booktabs,array,tabularx,multirow}
\usepackage[table]{xcolor}
\usepackage{float}
\usepackage{enumitem,pifont,xspace,url}
\usepackage[font=small]{caption}
\usepackage[colorlinks=true,citecolor=blue,linkcolor=blue,urlcolor=blue]{hyperref}
\usepackage[most]{tcolorbox}
\hypersetup{pdftitle={VideoLoop: Looped Working Memory Against Semantic Thrashing in Long-Form Video Agents},
  pdfauthor={Jianming Xu, Jinfa Huang, Jingyang Lin, Zhengyuan Yang, Jiebo Luo}}
\definecolor{TitleNavy}{HTML}{1B2A5C}
\definecolor{LinkNavy}{HTML}{1F3A8A}
\hypersetup{citecolor=LinkNavy, linkcolor=LinkNavy, urlcolor=LinkNavy}
\definecolor{AbstractBg}{HTML}{F5F7FA}
\fancypagestyle{firstpage}{\fancyhf{}\renewcommand{\headrulewidth}{0pt}\cfoot{\thepage}}
\newcommand{\method}{VideoLoop\xspace}
\newcommand{\myparagraph}[1]{\paragraph{#1}}
\newcommand{\cmark}{\ding{51}}

\definecolor{LightGreen}{RGB}{220,240,220}
\definecolor{LightBlue}{RGB}{220,230,245}
\definecolor{AliceBlue}{HTML}{F0F8FF}
\definecolor{LightGray}{RGB}{240,240,240}
\newcommand{\calD}{\mathcal{D}}
\newcommand{\calF}{\mathcal{F}}
\newcommand{\calH}{\mathcal{H}}
\newcommand{\calI}{\mathcal{I}}
\newcommand{\calK}{\mathcal{K}}
\newcommand{\calM}{\mathcal{M}}
\newcommand{\calT}{\mathcal{T}}
\newcommand{\calV}{\mathcal{V}}

\providecommand{\DatasetVideoMMMUVideos}{300}
\providecommand{\DatasetVideoMMMUQuestions}{900}
\providecommand{\DatasetVideoMMMUQuestionsPerTrack}{300}
\providecommand{\ResNativeVideoMME}{80.7}
\providecommand{\ResAppendVideoMME}{81.9}
\providecommand{\ResRewriteVideoMME}{83.3}
\providecommand{\ResFullVideoMME}{85.8}
\providecommand{\AblationNativeQOne}{93.3}
\providecommand{\AblationNativeQTwo}{86.2}
\providecommand{\AblationNativeQThree}{73.3}
\providecommand{\AblationNativeQFour}{69.8}
\providecommand{\AblationAppendQOne}{93.8}
\providecommand{\AblationAppendQTwo}{83.6}
\providecommand{\AblationAppendQThree}{79.6}
\providecommand{\AblationAppendQFour}{70.7}
\providecommand{\AblationRewriteQOne}{94.7}
\providecommand{\AblationRewriteQTwo}{86.7}
\providecommand{\AblationRewriteQThree}{79.6}
\providecommand{\AblationRewriteQFour}{72.4}
\providecommand{\AblationFullQOne}{94.7}
\providecommand{\AblationFullQTwo}{87.1}
\providecommand{\AblationFullQThree}{81.3}
\providecommand{\AblationFullQFour}{80.0}
\providecommand{\AblationAppendVsNativeGain}{1.2}
\providecommand{\ResFullVsNativeGain}{5.1}
\providecommand{\AblationRewriteVsAppendGain}{1.4}
\providecommand{\AblationRewriteVsAppendQTwoGain}{3.1}
\providecommand{\ResFullVsAppendGain}{3.9}
\providecommand{\AblationFullVsAppendQFourGain}{9.3}
\providecommand{\ResFullVsRewriteGain}{2.4}
\providecommand{\AblationFullVsRewriteQThreeGain}{1.8}
\providecommand{\AblationFullVsRewriteQFourGain}{7.6}
\providecommand{\TokenAppendVideoMME}{81.9}
\providecommand{\TokenRewriteVideoMME}{83.3}
\providecommand{\TokenFullVideoMME}{85.8}
\providecommand{\TokenFullVsAppendGain}{3.9}
\providecommand{\VLProVideoMME}{88.3}
\providecommand{\GainProVideoMME}{4.5}
\providecommand{\VLProVideoMMMU}{88.8}
\providecommand{\GainProVideoMMMU}{4.2}
\providecommand{\VLProLongVideoBench}{80.9}
\providecommand{\GainProLongVideoBench}{3.2}
\providecommand{\GainFlashVideoMME}{5.1}
\providecommand{\AbstractBackboneMeanGain}{4.2}
\providecommand{\GainProVsBestAgentVideoMME}{7.1}
\providecommand{\GainProVsBestAgentVideoMMMU}{10.4}
\providecommand{\GainProVsBestAgentLongVideoBench}{4.5}

\iclrfinalcopy
\begin{document}
\lhead{Preprint}
\thispagestyle{firstpage}
\begingroup
\setlength{\parindent}{0pt}
\vspace*{-0.42in}
\includegraphics[height=1.35cm]{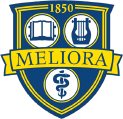}\hspace{6pt}\includegraphics[height=1.35cm]{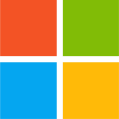}\par
\vspace{8pt}
{\color{TitleNavy}\rule{\textwidth}{1pt}}\par
\vspace{8pt}
{\centering\bfseries\LARGE VideoLoop: Looped Working Memory Against\\[2pt]
Semantic Thrashing in Long-Form Video Agents\par}
\vspace{7pt}
{\color{TitleNavy}\rule{\textwidth}{1pt}}\par
\vspace{14pt}
{\centering
{\bfseries Jianming Xu$^{1*}$, Jinfa Huang$^{1*}$, Jingyang Lin$^{1}$, Zhengyuan Yang$^{2}$, Jiebo Luo$^{1}$\par}
\vspace{2pt}
$^{1}$University of Rochester, $^{2}$Microsoft\qquad $^{*}$Equal contribution\par}
\vspace{13pt}
\begin{tcolorbox}[enhanced, colback=AbstractBg, colframe=TitleNavy, boxrule=0.8pt, arc=6pt,
  left=19.5pt, right=19.5pt, top=4pt, bottom=6pt]
{\centering\large\sc\color{TitleNavy}Abstract\par}
\vspace{3pt}
Long-form video understanding requires multimodal agents to iteratively gather evidence over many reasoning steps. However, most existing agentic methods suffer from \emph{semantic thrashing}: as append-only working memory grows, attention to key evidence collapses, and the agent loses access to what it has already found. First, we provide a structural argument showing that append-only memory can incorporate newly observed target evidence, but cannot remove accumulated
noise or prevent ordered context growth without a rewrite operator. Second, motivated by this analysis, we propose \textbf{\method{}}, a multimodal agent with two coupled loops. The outer loop reasons over the video and the inner loop, after each step, retrieves artifacts from an unbounded filesystem of past observations and intermediate analysis, and rewrites a bounded working memory. Extensive experiments demonstrate the effectiveness of VideoLoop, which improves four popular LVLM backbones in a plug-and-play manner, with an average gain of \AbstractBackboneMeanGain{}\% points over baseline on VideoMME (\emph{long}).
Further analysis of working memory suggests that VideoLoop mitigates semantic thrashing: on the hardest quarter of VideoMME (\emph{long}) questions, a blind judge that reads only the agent's context answers 81.1\% correctly, versus 60.9\% for the append-only agent.
  With Gemini 3.1 Pro, VideoLoop reaches \VLProVideoMME\% on VideoMME (\emph{long}), \VLProVideoMMMU\% on VideoMMMU, and \VLProLongVideoBench\% on LongVideoBench (\emph{long}).
\par\vspace{5pt}
{\small\textbf{Email:} \href{mailto:jhuang90@cs.rochester.edu,jluo@cs.rochester.edu}{\texttt{\{jhuang90, jluo\}@cs.rochester.edu}}, \href{mailto:philipxjm1@gmail.com}{\texttt{philipxjm1@gmail.com}}\par
\textbf{Code:} \href{https://github.com/philipxjm/videoloop}{\nolinkurl{github.com/philipxjm/videoloop}}\par}
\end{tcolorbox}
\endgroup
\vspace{-8pt}
\section{Introduction}
\label{sec:introduction}

Long-form video understanding~\citep{lin2026videoseek,luo2024video,wang2025videotree,tang2025video} requires reasoning over thousands of frames spanning minutes to hours, where the evidence relevant to a question is sparse and scattered across non-contiguous segments. Although recent studies attempt to scale representations to hour-level contexts~\citep{lin2025unleashing,shu2025video}, untrimmed videos contain massive natural redundancy~\citep{yao2025timechat,li2025empirical}. Therefore, numerous methods explore adaptive temporal search, pivot frame retrieval, and step-by-step reasoning~\citep{ye2025re,gao2026apvr,li2025divide,bhatnagar2026videomind,li2026kfs}. In this setting, general large vision-language models (LVLMs)~\citep{team2023gemini,team2024gemini,googledeepmind2026gemini31pro,openai2023gpt,openai2024gpt4o} still struggle to ingest such massive contexts reliably. Agentic systems~\citep{li2026lenswalk,lin2026videoseek,yan2026symphony,zhang2025deep,he2025framethinker} address this gap by reasoning iteratively: at each step, the agent observes a region of the video, updates a working memory of accumulated evidence, and decides where to look next. Iterative agents consistently outperform baselines on long-form video benchmarks.

\begin{figure}[!t]
\centering
\includegraphics[width=\linewidth]{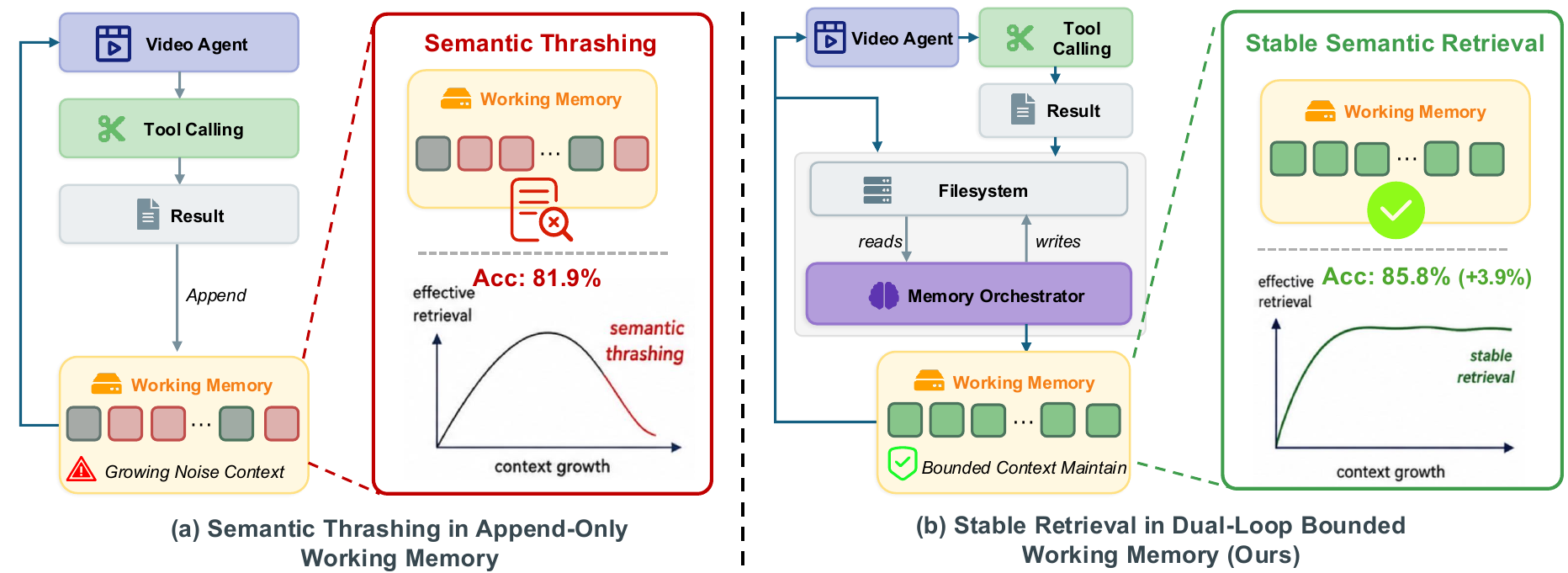}
\caption{\textbf{Existing Append-only working memory leads to semantic thrashing, our \method{} sustains stable retrieval via a dual-loop bounded working memory.}
\textbf{(a)} A single-loop video agent appends every tool result to a working memory bounded by the LVLM context. As context accumulates, attention to key evidence is diluted, and effective retrieval first peaks and then collapses as context continues to grow. The append-only agent achieves an \TokenAppendVideoMME\% final accuracy on VideoMME (\emph{long}).
\textbf{(b)} VideoLoop adds an inner orchestration loop. Tool results are written to an unbounded filesystem, and a memory orchestrator retrieves from the filesystem to rewrite a bounded working memory after every outer step. Our method improves task accuracy \mbox{by +\TokenFullVsAppendGain{} points on VideoMME (\emph{long}) to \TokenFullVideoMME\%.} Both configurations use Gemini 3 Flash.}
\label{fig:videoloop_framework}
\vspace{-4mm}
\end{figure}

However, most prior agentic methods~\citep{zhang2025deep,li2026lenswalk,yan2026symphony,he2025framethinker,wang2024videoagent} share a restrictive design: a single reasoning loop with an \emph{append-only} working memory. As the agent observes more, this memory accumulates within the LVLM's bounded context. Once it exceeds this bound, attention to key evidence of the user's query is severely diluted by the accumulation of observations, and the agent loses reliable access to what it has already found. As shown in Figure~\ref{fig:videoloop_framework}(a), we describe this systemic failure as \emph{semantic thrashing}, in analogy to OS thrashing~\citep{denning1968thrashing}: the agent expends increasing computational effort while its grounding on prior findings degrades, as a thrashing OS spends most cycles swapping pages \mbox{rather than running useful work.} We further show that semantic thrashing is a structural property of append-only updates rather than an incidental tuning issue. Treating the working memory and the optimal evidence set as subsets of a common evidence universe, append-only can reduce missing target evidence when new relevant observations arrive,
but cannot remove accumulated irrelevant evidence once it has entered memory. Closing this gap requires an operator that can \emph{remove} evidence from the working memory, which append-only memory by definition forbids.
Moreover, append-only memory grows increasingly costly as the ordered observation log expands. Therefore, append-only memory fails along two dimensions: it cannot discard irrelevant evidence, and it cannot bound the cost of retaining past observations.

These limitations motivate a decoupled design with two complementary properties: (i) an unbounded external store retaining all observations losslessly, and (ii) a bounded working memory rewritten at every step under a fixed budget. To this end, in Figure~\ref{fig:videoloop_framework}(b), we propose \textbf{\method{}}, a dual-loop architecture,  which realizes this decoupled design. The \emph{outer loop} is a multimodal agent that observes the video and writes its findings to a persistent filesystem of past observations and intermediate analysis. The \emph{inner loop} is a memory orchestrator that, after each outer step, retrieves question-relevant artifacts from the filesystem and rewrites a concise working memory. Decoupling reasoning from memory management allows each loop to operate within a fixed context budget, while the orchestrator retains random-access read privileges on the full trajectory history during consolidation. Furthermore, we evaluate our VideoLoop on three popular benchmarks for long-form video understanding. To probe memory quality, a blind judge that reads only the agent's accumulated context shows that append-only retrievability drops from 86.7\% on the easiest tasks (Q1) to 60.9\% on the hardest (Q4), while our method drops only from 94.7\% to 81.1\%. With Gemini 3.1 Pro as the policy model, it reaches \VLProVideoMME\% on VideoMME (\emph{long}), \VLProVideoMMMU\% on VideoMMMU, and \VLProLongVideoBench\% on LongVideoBench (\emph{long}), with absolute gains of 3.2\% to 4.5\% over the native LVLM. The dual-loop architecture is also plug-and-play across LVLM backbones, yielding consistent improvements without retraining. Overall, \mbox{our contributions are summarized as follows:}

\begin{itemize}[leftmargin=*,itemsep=2pt]
\item \textbf{Structural motivation for semantic thrashing.} We formalize semantic thrashing as a structural failure mode of append-only video-agent memory: such updates fail to
remove accumulated irrelevant evidence or prevent ordered context growth without a rewrite operator.

\item \textbf{Dual-loop bounded working memory.} VideoLoop decouples reasoning from memory management through an unbounded filesystem and bounded working memory rewritten at every step.

\item \textbf{Strong empirical performance.} Extensive experiments demonstrate that VideoLoop effectively mitigates memory degradation, improves accuracy on three long-form video benchmarks, and generalizes across diverse LVLM backbones without retraining.
\end{itemize}

\section{Related Work}
\label{sec:related_work}

\myparagraph{Agentic Video Understanding.}
Agentic video understanding pairs an LLM controller with a set of multimodal tools inside an iterative control loop~\citep{yao2023react,zhang2025deep,wang2024videoagent,fan2024videoagent,zhang2024simple,wang2025videotree,pang2025mr,lin2026videoseek,li2026lenswalk,zhang2024omagent,yang2025vca,chen2025lvagent,rege2026agentic,yan2026symphony,liu2025longvideoagent,yu2026longvidsearch}. At each step the controller reads the current state, picks a tool, and adds the result to what it knows about the video. The hard part is gathering and integrating evidence across temporal spans longer than any single context window. Most existing systems handle this with predefined pipelines or prebuilt indices~\citep{zhang2024simple,pang2025mr,chen2025lvagent,yan2026symphony,wang2024videoagent,wang2025videotree,lin2026videoseek,li2026lenswalk}, which are predictable but rigid. ReAct-style observe--act--reason loops~\citep{yao2023react} are the most common design, applied to keyframe selection, caption chains, and tree-structured retrieval. DVD~\citep{zhang2025deep}, for example, builds a multi-granular video database that a ReAct-style agent queries. Differently, our VideoLoop runs inside a coding sandbox and writes code to direct its own exploration. With only a minimal toolset, it composes retrieval and analysis routines as it goes, without upfront video preprocessing or committing to a fixed schema.

\myparagraph{Memory Mechanisms in Video Agents.}
Agentic memory has become a common way to handle long-context reasoning, with applications in long-form video understanding~\citep{yin2026videoarm,long2026seeing,yeo2026worldmm,wang2025video,hu2025hiagent,lin2023videoxum,wang2023lifelongmemory,lv2026all,sanders2024grounding} and persistent dialogue agents~\citep{packer2023memgpt,zhong2024memorybank,chhikara2025mem0,zhou2023recurrentgpt,feng2026m2a}. One family compresses memory through sliding windows, KV-cache pruning, and backtracking~\citep{xie2025video,yang2025cambrian,zuo2025videolucy} or learned retention and deletion in VideoMem's global memory buffer~\citep{jin2025videomem}. VideoARM records observations and reasoning traces in hierarchical multimodal memory~\citep{yin2026videoarm}. WorldMM uses an LLM to consolidate semantic triplets into an evolving knowledge graph~\citep{yeo2026worldmm}, while MemGPT supports active context management for general-purpose agents~\citep{packer2023memgpt}. In contrast, VideoLoop decouples video exploration from working-memory maintenance through a dedicated inner-loop agent. After each outer step, this agent combines three capabilities: (i) active retrieval of prior evidence from a persistent filesystem, (ii) joint consolidation of evidence across video segments and memory sections, and (iii) section-level memory editing under a fixed token budget.

\section{Methodology}
\label{sec:method}

\subsection{Semantic Thrashing Problem}
\label{sec:problem_formulation}

To formally analyze the failure modes of long-form video agents, we establish a structural analogy between the effective context limits of large vision-language models (LVLMs)~\citep{team2023gemini,team2024gemini,googledeepmind2026gemini31pro,openai2023gpt,openai2024gpt4o} and the memory-scheduling limits of classical operating systems (OS).

\textbf{Standard OS Thrashing}~\citep{denning1968thrashing}.
In multiprogramming environments, a process's active memory demand can be characterized by its \emph{working set}~\citep{denning1968working}.
In particular, the working set $W_i(t,\delta)$ denotes the set of distinct memory pages referenced by process $i$ within the recent time window $[t-\delta,t]$.
Let $M$ denote the total physical memory capacity, and let $\mathcal{I}(t)$ be the set of active processes at time $t$.
\emph{Thrashing}~\citep{denning1968thrashing} occurs when aggregate memory demand $\Lambda(t)$ exceeds physical capacity $M$:
\begin{equation}
\Lambda(t) \;=\; \sum_{i\in\mathcal{I}(t)} \big|\, W_i(t,\delta) \,\big| \;>\; M.
\label{eq:os_thrashing}
\end{equation}
Once this persists, the OS spends most cycles swapping pages, and useful throughput collapses.

\textbf{Semantic Thrashing in Video Agents.}
We formalize long-form video understanding as a sequential process of evidence gathering.
Given a video $\calV$ and query $q$, let $\mathcal{U}_q$ denote the universe of atomic evidence units relevant to query $q$, and let $\calK_q\subseteq\mathcal{U}_q$ denote the latent set of critical evidence required to answer $q$.
At reasoning step $t$, the agent ingests a new observation $o_t$ and updates its working memory $\calM_t$.
Most prior video agentic systems~\citep{he2025framethinker,li2026lenswalk,lin2026videoseek,zhang2025deep} update the working memory $\calM_t$ in an \emph{append-only} manner:
\begin{equation}
\calM_t \;=\; \calM_{t-1} \cup \{o_t\}.
\label{eq:append_only}
\end{equation}
However, current LLMs/LVLMs exhibit a bounded effective context capacity $C$~\citep{an2024does,hsieh2024ruler,liu2024lost}.
As working memory $\calM_t$ grows beyond this effective bound, $|\calM_t| > C$, the agent can no longer reliably retrieve and integrate the evidence in $\calK_q$.
Consequently, additional observations can dilute relevant evidence and reduce the reliability of retrieval and reasoning.
Analogous to OS thrashing, we term this failure mode \emph{semantic thrashing}: the agent keeps accumulating evidence while its reasoning grows unstable and less grounded.

\myparagraph{State Divergence as A Structural Indicator of Semantic Thrashing.}
We use the following as a conceptual diagnostic rather than a theorem-like reduction from OS thrashing. 
Treating $\calM_t$ and $\calK_q$ as subsets of a common evidence universe, we decompose their gap into \emph{missing target evidence} $\Delta_{\text{target}}^{(t)} = \calK_q \setminus \calM_t$ and \emph{redundant noise} $\Delta_{\text{pred}}^{(t)} = \calM_t \setminus \calK_q$, and define the \emph{state divergence}:
\begin{equation}
\mathrm{Dist}(\calM_t, \calK_q)
\;:=\;
\big|\Delta_{\text{target}}^{(t)}\big| + \big|\Delta_{\text{pred}}^{(t)}\big|
\;=\;
\big|\, \calM_t \,\triangle\, \calK_q \,\big|.
\label{eq:dist}
\end{equation}
This symmetric-difference cardinality is a conceptual diagnostic for memory quality.

With $M_t=M_{t-1}\cup \{o_t\}$, one update changes the
diagnostic by
\begin{equation}
{
\mathrm{Dist}(\calM_t, \calK_q) - \mathrm{Dist}(\calM_{t-1}, \calK_q)
=
-|\{o_t\}\cap(\calK_q\setminus \calM_{t-1})|
+|\{o_t\}\setminus(\calK_q\cup \calM_{t-1})|.
}
\label{eq:divergence}
\end{equation}
Eq.~\ref{eq:divergence} separates the effect of appending $o_t$ into useful and noisy additions. The term ${o_t}\cap(\calK_q\setminus\calM_{t-1})$ captures newly observed target evidence that was missing from memory, thereby reducing divergence. In contrast, ${o_t}\setminus(\calK_q\cup\calM_{t-1})$ captures newly introduced content that is neither target evidence nor previously stored, thereby increasing divergence. Append-only memory cannot remove such noise once added, so noisy trajectories gradually consume the context budget, obscure target evidence, and lead to semantic thrashing.

\myparagraph{Implications for Working Memory Design.}
Eq.~\ref{eq:append_only} and Eq.~\ref{eq:divergence} show the reason why append-only working memory is fragile in long-horizon or high-noise scenarios: critical evidence can enter $\calM_t$, but non-target content in $\calM_t\setminus\calK_q$ remains unless later updates can remove or rewrite it. This motivates a \emph{decoupled} memory architecture with two complementary properties: (i)~an \textit{unbounded external store} that retains all observations and intermediate artifacts losslessly, so that no evidence is lost prematurely, and (ii) a \textit{bounded working memory rewritten after each step} under a fixed token budget, which can delete redundant content in $\calM_t\setminus\calK_q$ and re-import \mbox{missing target evidence in $\calK_q\setminus\calM_t$ from~(i).}

\subsection{Video Agent with Persistent Sandbox}
\label{sec:agent}

\begin{figure}[t]
\centering
\includegraphics[width=\linewidth]{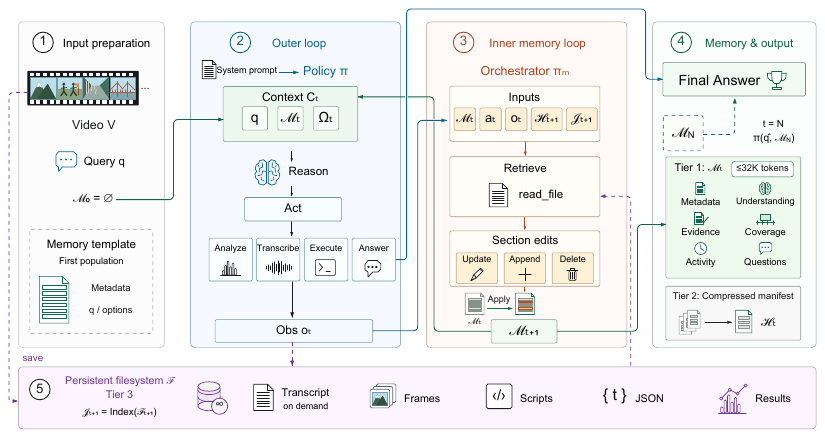}

\caption{\textbf{Overview of \method{}'s dual-loop architecture.} Working memory starts empty. The outer loop reasons about the task and explores the video with transcription and visual analysis tools as needed. After each outer step, the inner loop reads the question and the new observation, retrieves relevant evidence from the external store, and rewrites a compact working memory.}
\label{fig:dual_loop_method}

\end{figure}

\myparagraph{Overview.}
Figure~\ref{fig:dual_loop_method} illustrates the dual-loop architecture. Given a long-form video $\calV$ and a query $q$, \method{} operates as a sandboxed multimodal agent with five core components: a policy model $\pi$ instantiated by a multimodal LLM, a toolkit $\calT$, a working memory $\calM$, a memory orchestration $\pi_m$ for working memory rewriting, and a persistent sandbox environment with a unbounded filesystem $\calF$.
VideoLoop follows a dual-loop workflow: the outer loop employs the policy model $\pi$ to iteratively explore the video and save observations and artifacts to the filesystem $\calF$, while the inner loop uses $\pi_m$ to dynamically consolidate evidence units that can be recovered from those artifacts into an updated working memory.

\myparagraph{Basic Toolkit.}
The toolkit $\calT$ of the outer loop comprises four primitives:
\textsc{Analyze}($\hat{\calV}$, $\hat{q}$) answers the generated question $\hat{q}$ based on the selected video frames $\hat{\calV}$.
\textsc{Transcribe}$(\calV)$ returns the full timestamped transcript when requested by the policy and caches it after the first call.
\textsc{Execute}$(P)$ runs arbitrary code $P$ in a sandbox environment with image, video, and numerical libraries, persisting all artifacts (frames, captions, transcripts, a manifest of reasoning trace, etc.) on the sandbox filesystem $\calF$.
\textsc{Answer}$(y)$ commits an answer $y$ and terminates the agentic trajectory.

\subsection{Dual-Loop Bounded Working Memory}
\label{sec:dual_loop}

\myparagraph{Starting State.}
Working memory starts empty ($\calM_0 = \emptyset$), so at $t=0$ the outer policy sees only $q$. After the first step, $\pi_m$ writes the six-section template from $q$ and the first observation, including the video metadata, the question, and the options. The video is available in the sandbox, while the transcript is stored in the filesystem only after \textsc{Transcribe}$(\cdot)$ is called.

\myparagraph{Outer Loop: Agentic Reasoning \& Acting.}
At step $t$, $\pi$ reads the bounded context:
\begin{equation}
C_t \;=\; \{q, \calM_{t}, \Omega_t \},
\label{eq:context}
\end{equation}
where the $\calM_{t}$ denotes the current rewritten working memory, and $\Omega_t$ is a sliding window containing the thought-action-observation triplets from the nearest $k$ most recent iterations.
Conditioned on the context $C_t$, the policy model produces reasoning $r_t$ and an action $a_t$, and executing $a_t$ returns an observation $o_t$, respectively. The new observation $o_t$ will be written into the filesystem before its useful evidence is selectively admitted into $\calM_{t+1}$.
The outer loop \emph{never} ingests raw artifacts from prior iterations: it only accesses user query $q$, current work memory $\calM_{t}$, and the nearest sliding window $\Omega_t$.
Each component of $C_t$ is size-controlled: the working memory is bounded by a predefined token budget $B_{\calM}$, and $|\Omega_t|$ is capped at the $k$ most recent turns. Therefore, the total context size $|C_t|$ remains bounded by a \mbox{constant independent of the iteration $t$.}

\myparagraph{Inner Loop: Orchestrator Memory Rewrite.}
After each outer step, a separate LLM $\pi_m$, the \emph{memory orchestrator}, performs an active rewrite to produce a newly working memory $\calM_{t+1}$ based on the previous working memory $\calM_{t}$, the current action $a_t$ and the corresponding observation $o_t$, a compressed manifest $\calH_{t+1}$ of all prior actions, and $\calI_{t+1} = \text{Index}(\calF_{t+1})$ is an index over the current filesystem.
By design, $\pi_m$ guarantees the invariant $|\calM_t| \le B_{\calM}$ for all $t$, where $B_{\calM}$ is a token budget chosen below the effective context capacity $C$.

\myparagraph{Termination and Fallback Generation.}
The trajectory terminates whenever $a_t = \textsc{Answer}(\cdot)$, returning the final response $\hat{y}$. If the agent has not explicitly committed by the maximum iteration $N$, a fallback answer is generated by querying the policy directly on the last consolidated memory, $\hat{y} = \pi(\hat{q}, \calM_N)$, where $\hat{q}$ denotes the query $q$ appended with an instruction that requires the model to produce an answer immediately.

\subsection{Filesystem-based Memory Orchestration}
\label{sec:orchestrator}

Following the taxonomy of prior agentic systems~\citep{yin2026videoarm,jin2025videomem,li2026lenswalk}, we decompose $\pi_m$ along three axes: \emph{storage}, \emph{retrieval}, and \emph{consolidation}.

\myparagraph{Hierarchical Storage.}
VideoLoop maintains a three-tier memory hierarchy.
\textbf{Tier~1.} The bounded working memory $\calM_t$ visible to the outer loop is a typed document $\calM_t = \sigma_1^{(t)} \oplus \cdots \oplus \sigma_6^{(t)}$ with six ordered sections: metadata, narrative understanding (updated as needed), timestamped evidence, temporal coverage, activity log, and open investigation targets. 
The working memory $\calM_t$ is bounded by the token budget $B_{\calM}$.
\textbf{Tier~2.} The step manifest $\calH_t$ is a compressed log of all prior actions with their action parameters, such as timestamps, generated queries, and executed code. Entries older than the most recent $k$ are batch-summarized to govern which entries survive compression. 
\textbf{Tier~3.} The unbounded sandbox filesystem $\calF_t$ stores all extracted frames, analysis outputs, and intermediate scripts losslessly across iterations, preserving the raw material from which evidence units can be recovered.
$\calH_t$ bridges the filesystem and working memory \mbox{with a navigable, importance-weighted history.}

\myparagraph{Active Filesystem Retrieval.}
Before emitting its edits, $\pi_m$ retrieves question-relevant text artifacts, such as prior frame analyses and intermediate scripts, using $\calT_{\pi_m} = \{\texttt{read\_file}\}$. The filesystem index in its context lists the available files and frames.

\myparagraph{Working Memory Rewrite.}
At $t=0$, $\pi_m$ fills the empty memory with the six sections. At each later step, it emits edits $\calD_t \subseteq \{\textsc{Update},\allowbreak \textsc{Append},\allowbreak \textsc{Delete}\}$ for $\calM_t$, and sections without edits stay unchanged:
\begin{equation}
\calM_{t+1} \;=\; \textsc{Apply}\!\big(\,\calD_t,\; \calM_{t}\,\big), \qquad
\calD_t \;=\; \pi_m(q, \calM_{t}, a_t, o_t, \calH_{t+1}, \calI_{t+1}).
\label{eq:edits}
\end{equation}

The edits are per-section in syntax but \emph{cross-section in semantics}: the orchestrator $\pi_m$ decides $\calD_t$ over all sections jointly. For instance, the narrative understanding section $\sigma_2^{(t)}$ and the timestamped evidence section $\sigma_3^{(t)}$ are revised coherently against each other.

This closes the formal loop with Sec.~\ref{sec:problem_formulation}: when $\pi_m$ retrieves a missing target evidence item or discards redundant content, the rewrite is beneficial if removed noise plus imported target evidence outweighs deleted target evidence plus newly added noise. Let the following \mbox{four quantities summarize one rewrite:}\looseness=-1
\begin{equation}
\begin{aligned}
\rho_t &= |(\calM_t \setminus \calK_q) \setminus \calM_{t+1}|, \quad
\sigma_t = |(\calK_q \setminus \calM_t) \cap \calM_{t+1}|, \\
\mu_t &= |(\calK_q \cap \calM_t) \setminus \calM_{t+1}|, \quad
\nu_t = |\calM_{t+1} \setminus (\calK_q \cup \calM_t)|.
\end{aligned}
\end{equation}
Here $\rho_t$ is removed non-target content, $\sigma_t$ is imported missing target evidence, $\mu_t$ is target evidence accidentally removed, and $\nu_t$ is newly introduced non-target content. Then, we can derive
\begin{equation}
\label{eq:closure}
{
\mathrm{Dist}(\mathcal{M}_{t+1}, \mathcal{K}_q)
-
\mathrm{Dist}(\mathcal{M}_t, \mathcal{K}_q)
=
-\rho_t - \sigma_t + \mu_t + \nu_t.
}
\end{equation}
Therefore, $\mathrm{Dist}(\calM_{t+1},\calK_q)\le \mathrm{Dist}(\calM_t,\calK_q)$ only when removed noise and imported target evidence are at least as large as lost target evidence and newly added noise, \textit{i.e.}, $\rho_t+\sigma_t\ge \mu_t+\nu_t$, with strict improvement under strict inequality. This is a condition on rewrite quality, not an unconditional guarantee. Furthermore, active retrieval maintains a compact, query-relevant context across long-horizon trajectories, matching the design goal implied by Eq.~\ref{eq:closure}.

\section{Experiments}
\label{sec:experiments}

\begin{table}[!tb]
\centering
\footnotesize
\renewcommand{\arraystretch}{1.05}
\setlength{\tabcolsep}{3pt}
\resizebox{\linewidth}{!}{%
\begin{tabular}{@{}lcccccc@{}}
\toprule
\multirow{2}{*}{\textbf{Method}} & \multirow{2}{*}{\shortstack{\textbf{VideoMME}\\\textit{(long)}}} & \multicolumn{4}{c}{\textbf{VideoMMMU}} & \multirow{2}{*}{\shortstack{\textbf{LongVideoBench}\\\textit{(long)}}} \\
\cmidrule(lr){3-6}
& & \textbf{Per.} & \textbf{Comp.} & \textbf{Adapt.} & \textbf{Overall} & \\
\midrule
\multicolumn{7}{l}{\textit{Video agentic systems}} \\
VideoAgent~\citep{wang2024videoagent} & \makebox[6em][c]{49.0} & -- & -- & -- & \makebox[6em][c]{--} & \makebox[6em][c]{--} \\
LLoVi~\citep{zhang2024simple} & \makebox[6em][c]{50.2} & -- & -- & -- & \makebox[6em][c]{--} & \makebox[6em][c]{--} \\
VideoTree~\citep{wang2025videotree} & \makebox[6em][c]{54.2} & -- & -- & -- & \makebox[6em][c]{--} & \makebox[6em][c]{--} \\
QuoTA~\citep{luo2026quota} & \makebox[6em][c]{55.7} & -- & -- & -- & \makebox[6em][c]{--} & \makebox[6em][c]{--} \\
OmAgent~\citep{zhang2024omagent} & \makebox[6em][c]{60.5} & -- & -- & -- & \makebox[6em][c]{--} & \makebox[6em][c]{--} \\
VCA~\citep{yang2025vca} & \makebox[6em][c]{56.3} & -- & -- & -- & \makebox[6em][c]{--} & \makebox[6em][c]{--} \\
M3-Agent~\citep{long2026seeing} & \makebox[6em][c]{61.8} & -- & -- & -- & \makebox[6em][c]{--} & \makebox[6em][c]{--} \\
DVD~\citep{zhang2025deep} & \makebox[6em][c]{67.3} & -- & -- & -- & \makebox[6em][c]{--} & \makebox[6em][c]{68.6} \\
MR. Video~\citep{pang2025mr} & \makebox[6em][c]{63.4} & -- & -- & -- & \makebox[6em][c]{--} & \makebox[6em][c]{61.6} \\
LensWalk~\citep{li2026lenswalk} & \makebox[6em][c]{71.4} & 81.3 & 81.3 & 72.3 & \makebox[6em][c]{78.3} & \makebox[6em][c]{70.6} \\
VideoChat-A1~\citep{wang2026videochat} & \makebox[6em][c]{71.2} & -- & -- & -- & \makebox[6em][c]{--} & \makebox[6em][c]{--} \\
EGAgent~\citep{rege2026agentic} & \makebox[6em][c]{74.1} & -- & -- & -- & \makebox[6em][c]{--} & \makebox[6em][c]{--} \\
VideoSeek~\citep{lin2026videoseek} & \makebox[6em][c]{81.2} & -- & -- & -- & \makebox[6em][c]{--} & \makebox[6em][c]{73.5} \\
VideoARM~\citep{yin2026videoarm} & \makebox[6em][c]{81.2} & -- & -- & -- & \makebox[6em][c]{--} & \makebox[6em][c]{76.4} \\
\midrule
\multicolumn{7}{l}{\textit{Native LVLMs and Ours}} \\
GPT-4o~\citep{openai2024gpt4o} & \makebox[6em][c]{72.1} & 66.0 & 62.0 & 55.7 & \makebox[6em][c]{61.2} & \makebox[6em][c]{60.9} \\
Gemini 1.5 Pro~\citep{team2024gemini} & \makebox[6em][c]{77.4} & 59.0 & 53.3 & 49.3 & \makebox[6em][c]{53.9} & \makebox[6em][c]{58.6} \\
OpenAI o3~\citep{openai2025o3} & \makebox[6em][c]{84.9} & 79.3 & 75.7 & 71.3 & \makebox[6em][c]{75.4} & \makebox[6em][c]{--} \\
\addlinespace[2pt]
Gemini 3 Flash\textsuperscript{*}~\citep{googledeepmind2025gemini3flash} & \makebox[6em][c]{80.7} & 84.0 & 83.0 & 83.7 & \makebox[6em][c]{83.6} & \makebox[6em][c]{67.9} \\
\rowcolor{AliceBlue} \quad $+$\,\textbf{VideoLoop} & \makebox[6em][c]{85.8\rlap{\,{\scriptsize\textcolor{green!40!black}{(+5.1)}}}} & 90.3 & 87.3 & 85.3 & \makebox[6em][c]{87.7\rlap{\,{\scriptsize\textcolor{green!40!black}{(+4.1)}}}} & \makebox[6em][c]{73.8\rlap{\,{\scriptsize\textcolor{green!40!black}{(+5.9)}}}} \\
\addlinespace[2pt]
Gemini 3.1 Pro\textsuperscript{*}~\citep{googledeepmind2026gemini31pro} & \makebox[6em][c]{83.8} & 85.7 & 83.7 & 84.3 & \makebox[6em][c]{84.6} & \makebox[6em][c]{77.7} \\
\rowcolor{AliceBlue} \quad $+$\,\textbf{VideoLoop} & \makebox[6em][c]{\textbf{88.3}\rlap{\,{\scriptsize\textcolor{green!40!black}{(+4.5)}}}} & \textbf{90.7} & \textbf{87.7} & \textbf{88.0} & \makebox[6em][c]{\textbf{88.8}\rlap{\,{\scriptsize\textcolor{green!40!black}{(+4.2)}}}} & \makebox[6em][c]{\textbf{80.9}\rlap{\,{\scriptsize\textcolor{green!40!black}{(+3.2)}}}} \\
\bottomrule
\end{tabular}}
\caption{Benchmark comparison (accuracy, \%). Bold denotes column best; parentheses give overall gains over the preceding native backbone in percentage points. VideoMME and LongVideoBench use their long subsets with the official subtitles. VideoMMMU uses all 900 questions (300 per track): Perception (Per.), Comprehension (Comp.), Adaptation (Adapt.), and Overall. Results marked with \textsuperscript{*} are reproduced by us under the same settings. Dashes denote unavailable results.}
\label{tab:main_results}
\end{table}

\subsection{Experiment Setup}

\textbf{Evaluation Benchmarks}. We evaluate our \method{} on three video understanding benchmarks. 
VideoMME (\emph{long})~\citep{fu2025video} contains 900 multiple-choice questions over 30--60 minute videos, such as lectures, sports, documentaries, and entertainment. 
VideoMMMU~\citep{hu2025video} comprises \DatasetVideoMMMUQuestions{} questions drawn from \DatasetVideoMMMUVideos{} educational videos and evaluates models across the Perception, Comprehension, and Adaptation tracks, with \DatasetVideoMMMUQuestionsPerTrack{} questions per track. 
LongVideoBench (\emph{long})~\citep{wu2024longvideobench} evaluates detailed retrieval and reasoning across diverse web videos lasting up to 1 hour, with subtitles. We report results for its long split, comprising 564 questions from 188 videos, each 900--3600 seconds long.
Claude Opus 4.8 sorts VideoMME (\emph{long}) questions by difficulty into Q1--Q4 (easiest to hardest, 225 each), followed by human review.

\textbf{Baseline Methods}. We compare VideoLoop with native large vision-language models (LVLMs) that answer from video context in a single inference pass, and with video-agentic systems that perform multi-step video reasoning through iterative exploration.

\textbf{Implementation Details}.
We use Gemini 3.1 Pro and Gemini 3 Flash as the policy model for both the outer-loop agent and the inner-loop orchestrator.
For the \emph{outer-loop agent}, we set the maximum number of reasoning iterations $N$ to 50 and the minimum to 6. The \textsc{Analyze}($\cdot$) tool uses the default policy model in its native multimodal mode to inspect selected video content.
Working memory starts empty ($\calM_0 = \emptyset$), with no pre-loop initialization. The \textsc{Transcribe}($\cdot$) tool returns the official subtitles for VideoMME and LongVideoBench. VideoMMMU provides no subtitles, so we transcribe its audio with Whisper-large~\citep{radford2023robust}. The tool is invoked only when selected by the outer policy.
For the \emph{inner-loop orchestrator}, we set the working memory budget $B_{\calM}$ to 32K tokens and limited the number of selected key frames to a maximum of 6. In practice, we set the recent window to $k = 8$ message groups for the outer loop, and the orchestrator compresses the oldest $b = 10$ manifest records into one summary once $h = 30$ unsummarized records accumulate.

\subsection{Main Results}

\textbf{Overview.}
Table~\ref{tab:main_results} compares VideoLoop with native LVLMs and prior video agentic systems across three long-video benchmarks. With Gemini 3.1 Pro as the policy model $\pi$, VideoLoop obtains \VLProVideoMME\% on VideoMME (\emph{long}), \VLProVideoMMMU\% overall on VideoMMMU, and \VLProLongVideoBench\% on LongVideoBench (\emph{long}).

\textbf{Comparison with Base Models.}
Compared with its native base model, Gemini 3.1 Pro, VideoLoop improves by +\GainProVideoMME{}, +\GainProVideoMMMU{}, and +\GainProLongVideoBench{} points on VideoMME (\emph{long}), VideoMMMU, and LongVideoBench (\emph{long}), respectively. Its margins over the strongest prior agentic methods are +\GainProVsBestAgentVideoMME{}, +\GainProVsBestAgentVideoMMMU{}, and +\GainProVsBestAgentLongVideoBench{} points on the three benchmarks. The VideoMMMU breakdown shows improvements across all three cognitive tracks for both backbones.

\textbf{Effect of Policy Model $\pi$}.
Figure~\ref{fig:policy_model_ablation} evaluates different policy models as the reasoning engine.
VideoLoop consistently improves over the corresponding native LVLMs, achieving gains of $+4.5$, $+\GainFlashVideoMME$, $+3.5$, and $+3.7$ points with Gemini 3.1 Pro, Gemini 3 Flash, Kimi K2.5, and MiMo-V2-Omni, respectively.
These results suggest that VideoLoop is broadly compatible with different policy models and can effectively convert stronger reasoning engines into higher video understanding capability.\looseness=-1

\subsection{Analysis of VideoLoop Design Components}
\label{sec:ablation_videoloop}

\begin{figure*}[t]
\centering

\begin{minipage}[t]{0.48\textwidth}
\vspace{0pt}
\centering
\captionof{table}{Memory ablation on VideoMME (\emph{long}) with Gemini 3 Flash (one run per configuration). Accuracies use 900 questions overall and 225 per fixed-task-difficulty quartile (Q1 easiest, Q4 hardest), shared across rows. Gains use unrounded accuracies.}
\label{tab:memory_design_ablation}
\renewcommand{\arraystretch}{1}
\setlength{\tabcolsep}{5pt}
\resizebox{\linewidth}{!}{
\begin{tabular}{c c c c c c c}
\toprule
\multicolumn{2}{c}{\textbf{Memory Design}} 
& \multicolumn{5}{c}{\textbf{VideoMME (\emph{long})}} \\
\cmidrule(lr){1-2} \cmidrule(lr){3-7}
\textbf{\small Dual-Loop} 
& \textbf{\small Filesystem} 
& \textbf{\small All} 
& \textbf{\small Q1} 
& \textbf{\small Q2} 
& \textbf{\small Q3} 
& \textbf{\small Q4} \\
\midrule
  \multicolumn{2}{c}{\textit{Native Single-Pass LVLMs}} & \ResNativeVideoMME & \AblationNativeQOne & \AblationNativeQTwo & \AblationNativeQThree & \AblationNativeQFour \\
\midrule
\multicolumn{2}{c}{\textit{Append-only Agent}} & \ResAppendVideoMME & \AblationAppendQOne & \AblationAppendQTwo & \AblationAppendQThree & \AblationAppendQFour \\
{\cmark} & & \ResRewriteVideoMME & \textbf{\AblationRewriteQOne} & \AblationRewriteQTwo & \AblationRewriteQThree & \AblationRewriteQFour \\
\rowcolor{AliceBlue}
{\cmark} & {\cmark} & \textbf{\ResFullVideoMME} & \textbf{\AblationFullQOne} & \textbf{\AblationFullQTwo} & \textbf{\AblationFullQThree} & \textbf{\AblationFullQFour} \\
\bottomrule
\end{tabular}
}
\end{minipage}
\hfill
\begin{minipage}[t]{0.51\textwidth}
\vspace{0pt}
\centering
\captionof{figure}{Ablation study on policy model $\pi$.}
\includegraphics[width=\linewidth]{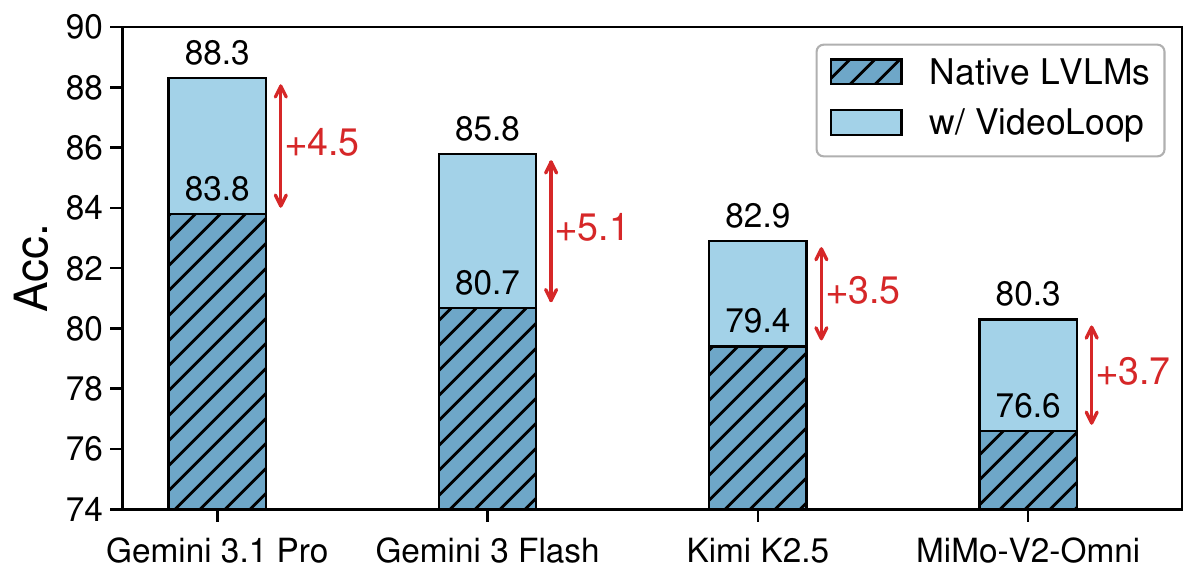}
\label{fig:policy_model_ablation}
\end{minipage}

\end{figure*}

Table~\ref{tab:memory_design_ablation} separates agentic exploration from memory design. Append-only raises accuracy over native inference from $\ResNativeVideoMME$ to $\ResAppendVideoMME$ ($+\AblationAppendVsNativeGain$). Dual-loop rewriting reaches $\ResRewriteVideoMME$ ($+\AblationRewriteVsAppendGain$ over append-only), with its largest gain on Q2 ($+\AblationRewriteVsAppendQTwoGain$) and no change on Q3. Adding filesystem access raises accuracy to $\ResFullVideoMME$ ($+\ResFullVsRewriteGain$ over dual-loop), gaining $+\AblationFullVsRewriteQThreeGain$ and $+\AblationFullVsRewriteQFourGain$ on Q3 and Q4. Full VideoLoop exceeds append-only by $+\ResFullVsAppendGain$ points (largest on Q4, $+\AblationFullVsAppendQFourGain$) and native inference by $+\ResFullVsNativeGain$ points.

\subsection{Further Analysis}

\begin{figure}[!tb]
\centering
\includegraphics[width=\linewidth]{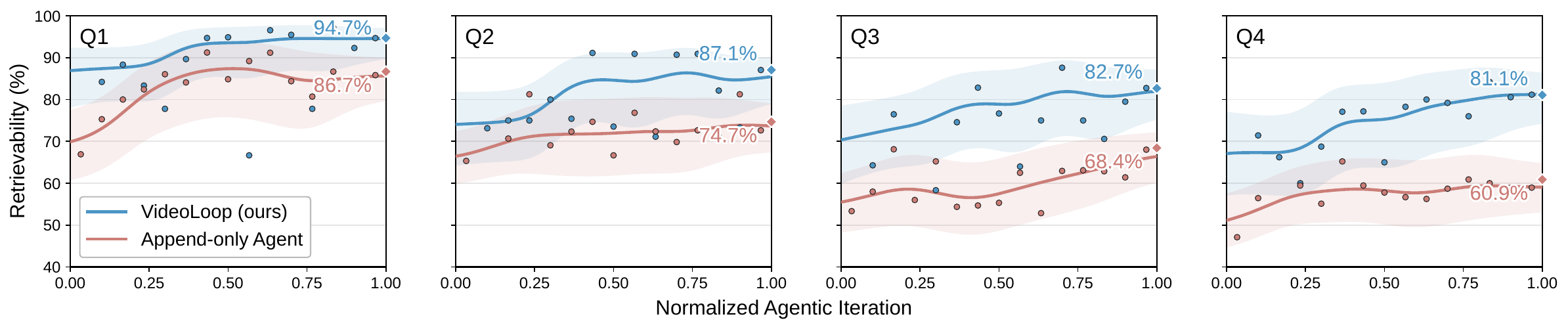}

\caption{Illustration of semantic thrashing on VideoMME (\emph{long}).
Four panels present fixed task difficulty quartiles Q1--Q4 from left to right (easiest to hardest), with the same grouping used across memory designs.
The curves show blind-judge answer accuracy on frozen context snapshots, used as a proxy for evidence retrievability. Diamonds and labels at the right end give final-snapshot accuracy.
Semantic thrashing occurs when retrievability early plateaus or \mbox{degrades despite continued agentic iterations.}}
\label{fig:thrashing_headline}

\end{figure}

\textbf{Evidence Retrievability Reveals Semantic Thrashing}.
Figure~\ref{fig:thrashing_headline} evaluates evidence retrievability through a context-only answerability proxy. At sampled iterations, a separate blind judge (Gemini 3.1 Flash-Lite) receives the question, answer options, and a frozen context snapshot, without access to the video, filesystem, or tools. Its answer accuracy measures how well the current context supports answering the question.
VideoLoop maintains stable or gradually improving retrievability across quartiles, showing that additional iterations help accumulate and preserve usable evidence.
In contrast, the append-only agent often plateaus early (Q2 and Q4) or even degrades (Q1), indicating that simply appending observations can dilute or obscure critical evidence rather than improve the effective working memory.
\mbox{As task difficulty} increases from Q2 to Q4, the gap becomes substantially larger: In Q4, the final-snapshot judge accuracy is 81.1\% for VideoLoop and 60.9\% for the append-only agent.

\begin{figure}[!tb]
\centering
\begin{minipage}{\linewidth}
  \captionof{table}{\textbf{Token cost vs. accuracy on VideoMME (\emph{long}).}                                                                         
  Per-question means, in thousands of tokens. Relative to append-only, full VideoLoop gains +\TokenFullVsAppendGain{} pp in accuracy with additional cost of 0.5\% tokens; dual-loop only gains +\AblationRewriteVsAppendGain{} pp with additional cost of 5.2\% tokens.}
  \label{tab:token_usage}
                                           
  \centering                                                                                                                                            
  \small                                                          
  \setlength\tabcolsep{8pt}                                                                                                                             
  \begin{tabularx}{\linewidth}{l*{5}{>{\raggedleft\arraybackslash}X}}                                  
  \toprule
  \multirow{2}{*}{\textbf{Memory Design}} & \multicolumn{3}{c}{\textbf{Tokens / Question (K)}} & \multicolumn{2}{c}{\textbf{Performance}} \\
  \cmidrule(lr){2-4} \cmidrule(lr){5-6}                                                                                                                 
  & \textbf{In} & \textbf{Out} & \textbf{Total} & $\bm{\Delta}$ & \textbf{Acc.} \\
  \midrule                                                                                                                                              
  (a) Append-only                       & 584.6 & 30.3 & 614.9          & ---             & \TokenAppendVideoMME \\
  (b) Dual-loop only                    & 590.0 & 57.0 & 647.0          & +5.2\%          & \TokenRewriteVideoMME \\                                                    
  \rowcolor{cyan!10}                                                                                                                                    
  (c) \textbf{VideoLoop} (\emph{ours})  & 559.0 & 59.2 & \textbf{618.2} & \textbf{+0.5\%} & \textbf{\TokenFullVideoMME} \\                                              
  \bottomrule                                                                                                                                           
  \end{tabularx}                                                                                                                                        
  \end{minipage}
\par\vspace{4pt}
\begin{minipage}{\linewidth}
\centering
\includegraphics[width=\linewidth]{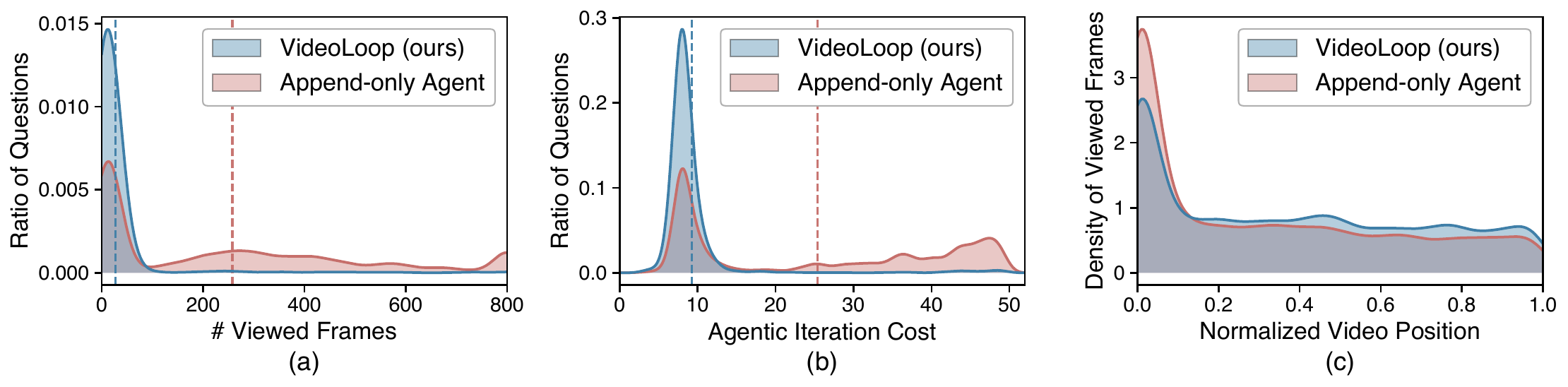}

\captionof{figure}{Agentic behavior analysis on VideoLoop and append-only agent on VideoMME (\emph{long}).
(a) Distribution of the number of viewed frames per question.
(b) Distribution of agentic iteration cost per question.
(c) KDE density of viewed frames over normalized video positions, where 0 and 1 indicate the beginning and end of a video.
Dashed lines in (a) and (b) indicate the mean values.}
\label{fig:behavior_analysis}
\end{minipage}

\end{figure}

\textbf{Token Efficiency Analysis across Memory Designs.}
Table~\ref{tab:token_usage} compares token usage and accuracy across different memory designs on VideoMME (\emph{long}) with Gemini 3 Flash.
The append-only baseline uses $614.9$K tokens per question and achieves $\TokenAppendVideoMME\%$ accuracy.
Adding the dual-loop workflow improves accuracy to $\TokenRewriteVideoMME\%$, but increases token usage to $647.0$K, a $+5.2\%$ overhead, mainly due to the additional output tokens introduced by memory rewriting (inner loop). 
In contrast, the full VideoLoop design with filesystem achieves the highest accuracy, $\TokenFullVideoMME\%$, while using only $618.2$K tokens, nearly matching the append-only baseline with just $+0.5\%$ overhead. 
Although VideoLoop produces more output tokens, it reduces the number of input tokens from $584.6$K to $559.0$K, suggesting that the filesystem helps externalize and selectively reuse intermediate evidence rather than repeatedly carrying the full history in context.

\textbf{Agentic Behavior Analysis}.
As shown in Figure~\ref{fig:behavior_analysis}(a), VideoLoop concentrates its viewed-frame distribution in the low-cost regime, while the append-only agent exhibits a substantially longer tail, indicating that append-only memory accumulation tends to trigger excessive visual inspection. 
A similar trend is observed in Figure~\ref{fig:behavior_analysis}(b): VideoLoop requires fewer agentic iterations per question, whereas the append-only agent often continues for many more reasoning steps.
These results suggest that \emph{VideoLoop's memory orchestration provides a more compact and effective working memory}, reducing redundant evidence collection and mitigating the iterative overhead caused by unstructured memory growth. 
Figure ~\ref{fig:behavior_analysis}(c) further indicates that \emph{VideoLoop produces a more temporally balanced distribution of viewed frames} across the normalized video timeline, rather than simply focusing on a narrow segment \mbox{at the beginning of the videos.}\looseness=-1

\section{Conclusion}
\label{sec:conclusion}

In this paper, we present \method{}, a dual-loop agentic framework for long-form video
understanding. Motivated by \emph{semantic thrashing} in previous video agents, where append-only context dilutes early evidence as the trajectory grows, we pair an outer multimodal agent with an inner LLM
that rewrites a bounded working memory after every outer step from a
sandboxed artifact store. \method{} improves accuracy on
three long-form video benchmarks, and the gain transfers across multiple
LVLM backbones in a plug-and-play manner. Beyond video, we see a general principle for long-horizon agents: memory should be curated, not accumulated.

\clearpage
\begingroup
\setlength{\bibsep}{2pt}
\bibliographystyle{iclr2027_conference}
\bibliography{ref}
\endgroup

\end{document}